\documentclass[letterpaper, 10 pt, conference]{ieeeconf}
\IEEEoverridecommandlockouts
\usepackage{epsfig}
\usepackage{times}
\usepackage{amsmath}
\usepackage{amssymb}
\usepackage{multirow}
\usepackage[T1]{fontenc}  
\usepackage[utf8]{inputenc}
\usepackage[english]{babel}
\usepackage[export]{adjustbox}
\usepackage{graphicx}
\usepackage{blindtext}
\usepackage{siunitx}
\usepackage{booktabs, threeparttable}
\usepackage{subcaption}
\usepackage[colorinlistoftodos, german]{todonotes}
\usepackage[T1]{fontenc}
\usepackage[utf8]{inputenc}
\usepackage{textcomp}
\usepackage[absolute,showboxes]{textpos}
\usepackage{hyperref}
\hypersetup{
    colorlinks=true,
    linkcolor=black,
    citecolor=black,
    filecolor=black,
    urlcolor=black,
pdfinfo={
Title={Limited Visibility and Uncertainty Aware Motion Planning for Automated Driving},
Author={Omer Sahin Tas},
Subject={IEEE Intelligent Vehicle Symposium 2018}
}
}
\TPMargin{5pt}
\newcommand{\copyrightstatement}{
    \begin{textblock}{0.84}(0.08,0.02)
         \noindent{\footnotesize{\copyright 2018 IEEE.
         Personal use of this material is permitted. Permission from IEEE must be obtained for all other uses, in any current or future media, including reprinting/republishing this material for advertising or promotional purposes, creating new collective works, for resale or redistribution to servers or lists, or reuse of any copyrighted component of this work in other works. DOI: \url{https://doi.org/10.1109/ITSC.2018.8569580}
         
         \vspace{2mm}
         \noindent
         In proceedings of the IEEE Intelligent Transportation Systems Conference (ITSC), Maui, Hawaii, USA, November 4-7, 2018, pp.\ 2419-2425}}
    \end{textblock}
}

\title{\LARGE \bf 
Decision-Time Postponing Motion Planning for Combinatorial Uncertain Maneuvering}

\author{
Ömer Şahin Taş,
Felix Hauser 
and Christoph Stiller
\thanks{
The authors are with 
FZI Research Center for Information Technology
at the Karlsruhe Institute of Technology, 
Haid-und-Neu-Str. 10-14, 76131 Karlsruhe, Germany.
E-mail: 
{\tt\small 
\{\texttt{tas}, 
\texttt{hauser},
\texttt{stiller\}@fzi.de}
}
}
}

\begin{document}
\copyrightstatement
\maketitle
\thispagestyle{empty}
\pagestyle{empty}

%%%%%%%%%%%%%%%%%%%%%%%%%%%%%%%%%%%%%%%%%%%%%%%%%%%%%%%%%%%%%%%%%%%%%%%%%%%%%%%%%%%%%%%%%%%%%%%%%%%%%%%%%%%%%%%%
\begin{abstract} 
Motion planning involves decision making among combinatorial maneuver variants in urban driving. 
A planner must consider uncertainties and associated risks of the maneuver variants, 
and subsequently select a maneuver alternative. 
In this paper we present a planning approach that considers the uncertainties in the prediction and, 
in case of high uncertainty, postpones the combinatorial decision making to a later time within the planning horizon. 
With our proposed approach, safe but at the same time not overconservative motion is planned.
\end{abstract}

%%%%%%%%%%%%%%%%%%%%%%%%%%%%%%%%%%%%%%%%%%%%%%%%%%%%%%%%%%%%%%%%%%%%%%%%%%%%%%%%%%%%%%%%%%%%%%%%%%%%%%%%%%%%%%%%
\section{Introduction}
\label{sec:introduction}

Motion planning for automated urban driving requires planning of overtaking and merge maneuvers. 
The presence of obstacles for the execution of those maneuvers introduce discrete solution classes. 
For an intersection scenario, a trivial example is either to drive straight or to yield the approaching vehicle. 
These discrete solution classes correspond to feasible, locally optimal motions.
Literature inspected this subject under 
\textit{homotopy classes}, 
\textit{semantic maneuver variants} or 
\textit{combinatorial alternatives}.   
If motion is planned with local search methods, 
the selection among distinct homotopic classes are typically done by divide-and-conquer strategy,
\textit{i.e.\ }by finding the local minimum in all classes and 
afterwards selecting the most optimal class.

Presented approaches in the literature make a decision among the variants at the time of planning. 
They typically do not consider the uncertainties resulting from prediction and 
the therefrom arising collision probabilities. 
Furthermore, they do not postpone the decision making to a later time with the 
expectation that the prediction would be more certain.
Only Markov Decision Process based methods intrinsically deal with this issue
by modeling the belief over the state space and and choosing behaviour to reduce uncertainty if it is too high.

In this paper we present a local-continuous planning approach that 
considers the uncertainties in the environment model by inspecting the resulting collision probabilities. 
The planner tries to achive the driving goals, while minimizing the associated collision probability. 
The planning is done compound for all of the homotopy classes: 
the distinct maneuver options are subject to the same cost functors and 
are solved in the same optimization problem. 
A parallel solver routine simultaneously considers to postpone the combinatorial decision making 
to a later time in the planning horizon. 
The planner finally either selects to execute one of the homotopy classes or 
continues to drive in a way that is optimal for both combinations and expects the prediction to be more certain in the future. 
With our proposed approach, combinatorial decision making is set on a probabilistic basis 
and safe motion without being overly conservative is obtained. 

\begin{figure}[t!]
\vspace{2mm}
\includegraphics[width=\columnwidth]{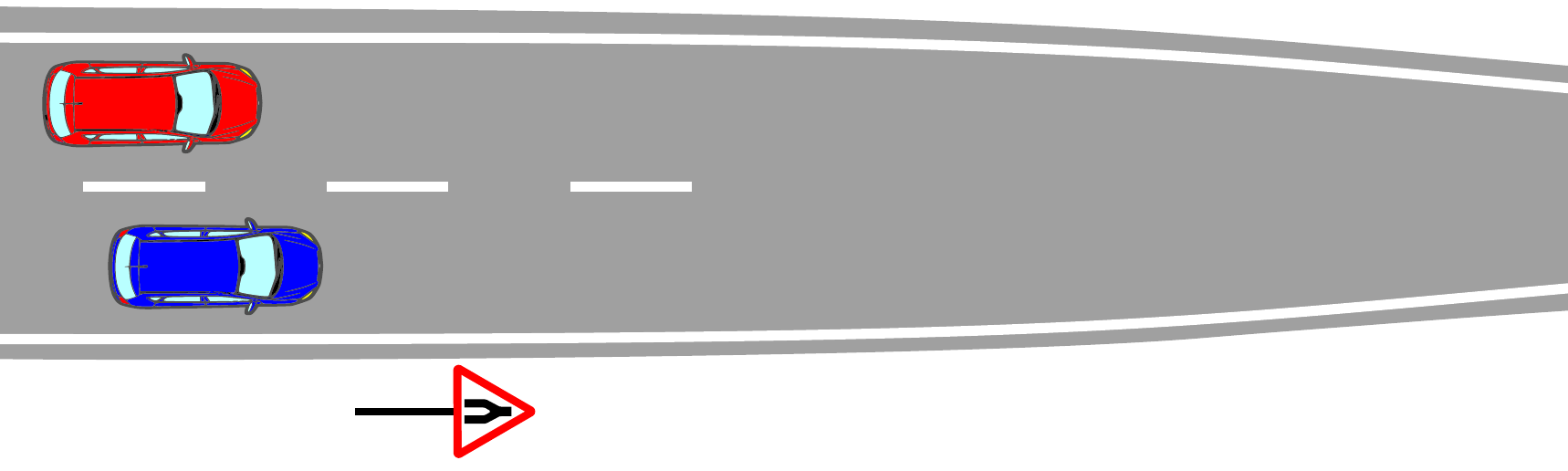}
\caption{A use case, where interaction and thus the estimation of other vehicle’s intention is required. 
If the other vehicle's intention is unclear, 
an optimal motion would minimize the total cost of both manuever alternatives.}
\label{fig:figure_01}
\end{figure}

The rest of the paper is structured as follows: 
In Section \ref{sec:related_work} we presented related work.
Then, in Section \ref{sec:environment_model}, 
we present the environment model and introduce the uncertainties in the prediction.  
The underlying motion prediction model and resulting probability distributions are also defined in this section. 
We afterwards continue with Section~\ref{sec:planning} and present our planning approach, \textit{c.f.\ }Fig.~\ref{fig:figure_01}.
Then in Section \ref{sec:experiments} we present the simulation results on a use case, 
where the utilization of this planning approach is advantageous. 
Finally, in Section \ref{sec:conclusion} we conclude the paper and summarize the key aspects.

%%%%%%%%%%%%%%%%%%%%%%%%%%%%%%%%%%%%%%%%%%%%%%%%%%%%%%%%%%%%%%%%%%%%%%%%%%%%%%%%%%%%%%%%%%%%%%%%%%%%%%%%%%%%%%%%
\section{Related Work}
\label{sec:related_work}

The combinatorial nature of motion planning in urban environments 
have been identified by several works at about the same time. 
The work in \cite{tas2014integrating} and \cite{bender2015combinatorial}
decompose the configuration space to find homotopic classes for a given environment model.
The former focuses on intersection crossing, whereas the latter focuses on overtaking. 
A similar work presented in \cite{kohlhaas2014semantic} 
distinguishes between semantic maneuver options in traffic.
Given the potential relations to the ego vehicle, 
semantic maneuvers are calculated on Lanelets. 

Partitioning of the configuration space into homotopic manuever classes 
can be quite burdensome. 
Besides the works \cite{tas2014integrating, bender2015combinatorial, kohlhaas2014semantic} 
that use specific obstacle representations for identifying maneuver options, 
another work presented in \cite{park2015homotopy} utilizes cell-decomposition methods. 
A very recent work \cite{altche2017partitioning} efficiently partitions 
collision-free configurations into sub-regions.
Another work \cite{sontges2017computing} proposes 
to use reachable sets and to filter out invalid combinations by 
utilizing a simple collision checker.

Partially Observable MDP (POMDP) methods 
take the effect of uncertainty on decision making into account
by considering the current belief over the state space. 
The resulting discrete action sequences typically result in risk-aware maneuver planning. 
The work \cite{brechtel2014probabilistic} 
considers various uncertainties in the motion planning including 
occluded regions and is further capable to run online.
On the other hand, it requires to train the scenes. 
Another POMDP framework presented in \cite{hubmann2017decision} does not require scene-specific training. 
The planner considers possible routes of other vehicles, 
as well as uncertainties in the perception and motion execution. 
The planner can postpone its decision among the maneuver options
by keeping track of changes in the belief. 
The work \cite{sefati2017towards} also utilizes a POMDP for longitudinal motion planning.
Uncertainties in the route, state and motion execution of the vehicles are considered as well.
The planner chooses between three different accelerations every time step, 
resulting in a motion that is quite jerky.
An application of reinforcement learning on a roundabout scenario is presented in \cite{gritschneder2016adaptive}.
They use a MDP to choose from high-level behavior actions (\textit{approach}, \textit{stop}, \textit{go} and \textit{uncertain}), which are then sent to the trajectory planner. 
The planner, in case, can postpone the decision making to a later time as well. 
However, the work neglects the uncertainty in state and route of the other vehicle.
A later work of the same research institution \cite{hoermann2017entering} 
predicts occupancy probabilities with a grid map and identifies free-to-drive sections of a path. 
With the free-to-drive sections, the planner can gradually approach an intersection. 
The work, nevertheless, does planning for a short horizon and hence remains reactive to other participants.

%%%%%%%%%%%%%%%%%%%%%%%%%%%%%%%%%%%%%%%%%%%%%%%%%%%%%%%%%%%%%%%%%%%%%%%%%%%%%%%%%%%%%%%%%%%%%%%%%%%%%%%%%%%%%%%%
\section{Environment Model}
\label{sec:environment_model}

The planner works on an environment model that contains all of the required input data. 
We assume that the road boundaries, lanes and hence the mid of the lanes are known in advance. 
The vehicles can only perceive other vehicles that are inside the visible area. 
The perception and prediction is bound to uncertainty. 
Whereas the route of the vehicles in the environment are assumed to be known, 
the maneuver intentions must be estimated. 

\subsection{Prediction Model}
\label{sec:idm}

An overview on motion prediction for automated vehicles is presented in \cite{lefevre2014survey}. 
We use the Intelligent Driver Model (IDM) \cite{treiber2000congested} as the underlying model 
for predicting the future motion of other vehicles. 
The IDM allows to model the interaction of a vehicle with other traffic participants, 
given model parameters such as set speed, safe speed, comfortable deceleration, and maximum acceleration. 

In our work, we assume the parameters of the IDM for predicting other vehicles is known beforehand. 
Hence, we neglect the prediction uncertainty arising from parameters of the underlying motion model.
We assume that other vehicles predict the motion of the ego vehicle with constant speed. 

\subsection{Uncertainty Propagation along Prediction Horizon}
\label{sec:uncertainty_propagation}

To model the Gaussian uncertainty propagation along the prediction horizon, 
we use a similar approach as presented in \cite{xu2014motion}. 
We model longitudinal motion of the vehicles with piece-wise constant accelerations and hence
define the system state $\mathbf{x}_{k} = [x,\, v,\, a]^\mathsf T$, 
the control input obtained from IDM as $\mathbf{u}_k = [a]$ 
and the measurement vector as $\mathbf{z}_k = [x,\, v]^\mathsf T$. 
We further assume the process noise $\mathbf Q_k$ to be a discrete time Wiener-process and hence
acceleration increments are independent. 
This yields the linear discrete time system
\begin{equation}
\mathbf{x}_{k} = \mathbf{F}_{k} \mathbf{x}_{k-1} + \mathbf{B}_{k} \mathbf{u}_{k} + \mathbf{w}_{k} \, , \hspace{2mm}  \mathbf{w}_k \sim \mathcal{N}(\mathbf{0}, \mathbf{Q}_k) \, .
\end{equation}
$\mathbf{F}_{k}$ is the state transition model,
$\mathbf{B}_{k}$ is the system input model,
$\mathbf{u}_{k}$ is the system input, and the
$\mathbf{w}_{k}$ is the process noise with covariance $\mathbf Q_k$. 
Likewise, we assume that the measurements $\mathbf{z}_k$ of the system state $\mathbf{x}_k$ at time step $k$ can be represented by
\begin{equation}
\mathbf{z}_k = \mathbf{H}_{k} \mathbf{x}_k + \mathbf{v}_k \, , \hspace{2mm} \mathbf{v}_k \sim \mathcal{N}(\mathbf{0}, \mathbf{R}_k) \, ,
\end{equation}
where 
$\mathbf{H}_{k}$ is the measurement model which maps the real state space into the measurement space, 
and 
$\mathbf{v}_k$ is the measurement noise with covariance $\mathbf{R}_k$.

The system matrices are given as
$$
\mathbf{F}_k = 
\begin{bmatrix}
1 & \Delta t & {\Delta t}^2/2 \\ 
0 & 1 & \Delta t\\ 
0& 0& 1
\end{bmatrix}
, \hspace{3mm}
\mathbf{Q}_k = 
\begin{bmatrix} 
\frac{\Delta t^4}{4} & \frac{\Delta t^3}{2} & \frac{\Delta t^2}{2}  \\ 
\frac{\Delta t^3}{2} & \Delta t^2 & \Delta t \\ 
\frac{\Delta t^2}{2}& \Delta t & 1
\end{bmatrix}
\sigma_a ^ 2
$$
 
$$
\mathbf B_k = 
\begin{bmatrix}
\frac{1}{2}\Delta t^2 \\ 
\Delta t \\ 
1
\end{bmatrix}
, \hspace{3mm}
\mathbf H_k = 
\begin{bmatrix}
1 & 0 & 0 \\ 
0 & 1 & 0
\end{bmatrix} 
\, ,
$$
where ${\Delta t}$ represents the sampling time 
and 
${\sigma_a ^ 2}$ the variance in process noise. 
The sampling frequency $\frac{1}{\Delta t}$ of the Kalman filter
is selected as multitude of new sensor measurement arrival time to the estimator.
For convenience, we set it the same as the sampling interval of the motion planner. 

For the given linear system and measurement model, 
a Kalman filter can be used for uncertainty calculation. 
The Kalman equations are given as:
\begin{align}
\text{Prediction Step} \nonumber \\
\hat{\mathbf x}_{k}^- &= \mathbf F_k\hat{\mathbf x}^{+}_{k-1} + \mathbf B_k \mathbf u_k \label{eq:kalman_input}  \\
\mathbf P_{k}^- &=  \mathbf F_k \mathbf P^{+}_{k-1} \mathbf F_k^\mathsf T + \mathbf Q_k  \\ 
\nonumber  \\       
\text{Update Step} \nonumber \\
\mathbf K_k &= \mathbf P_{k}^-\mathbf H_k^\mathsf T (\mathbf H_k \mathbf P_{k}^- \mathbf H_k^\mathsf T + \mathbf R_k)^{-1}\\
\hat{\mathbf x}_{k}^{+} &= \hat{\mathbf x}_{k}^- + \mathbf K_k (\mathbf z_k - \mathbf H_k\hat{\mathbf x}_{k}^-)\\
\mathbf P_{k}^{+} &= (I - \mathbf K_k \mathbf H_k) \mathbf P_{k}^- 
\end{align}
where 
$\hat{\mathbf x}_{k}^-$ is the prior estimate of the state,
$\hat{\mathbf x}_{k}^{+}$ is the posterior estimate of the state,
$\mathbf P_{k}^-$ is the prior error covariance,  
$\mathbf P_{k}^{+}$ is the posterior error covariance and reflects the accuracy of the state estimate,
and $\mathbf K_k $ is the Kalman gain.

\begin{figure}
\vspace{2mm}
    \centering
    
    \begin{subfigure}[b]{0.45\columnwidth}      
        \includegraphics[width=\columnwidth]{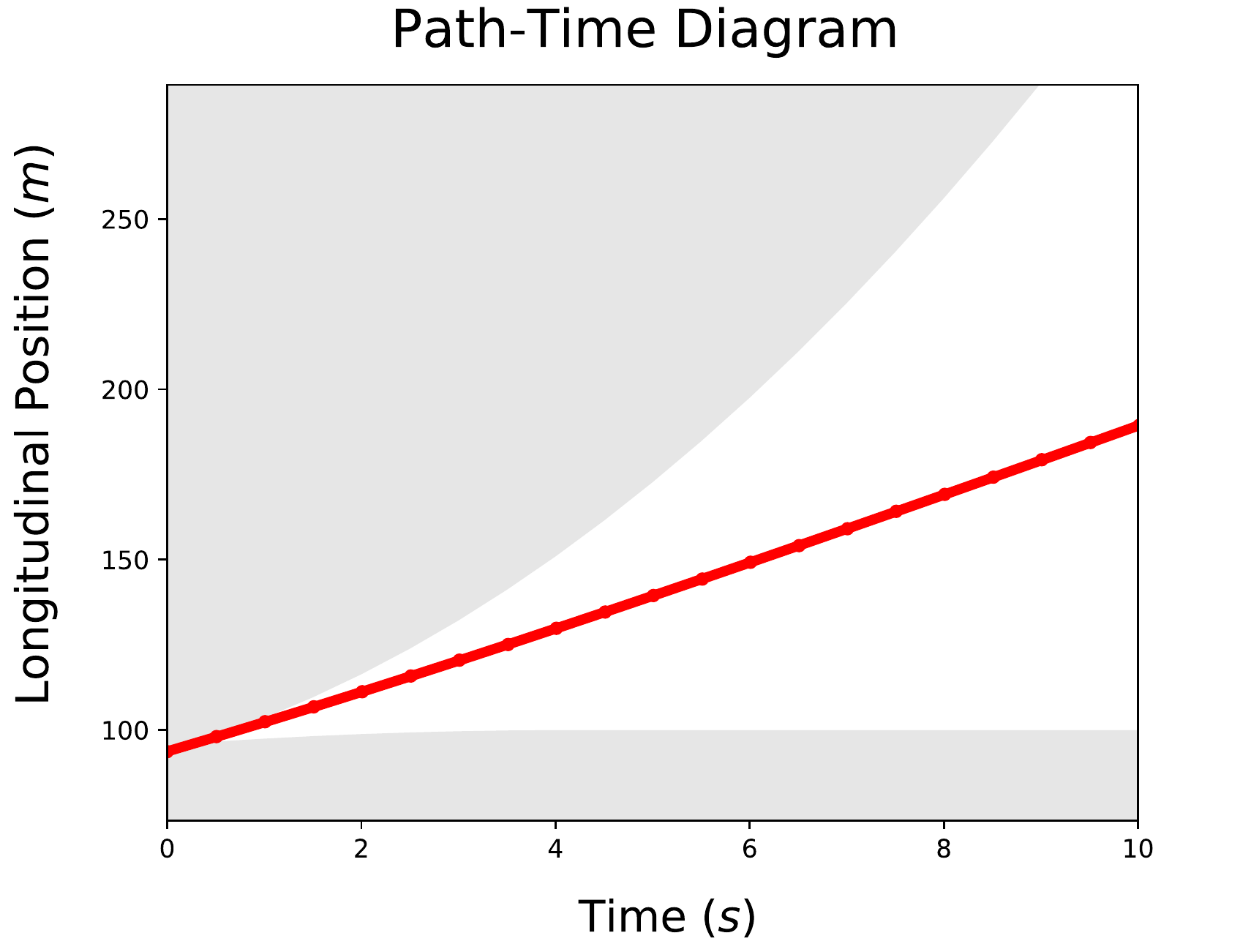}
        \caption{Predicted motion by using IDM on path-time diagram. 
        The grey filled areas represent the regions that lie out of the reachability of the vehicle.}
        \label{fig:figure_02a}
    \end{subfigure}\hfill
    \begin{subfigure}[b]{0.45\columnwidth}      
        \includegraphics[width=\columnwidth]{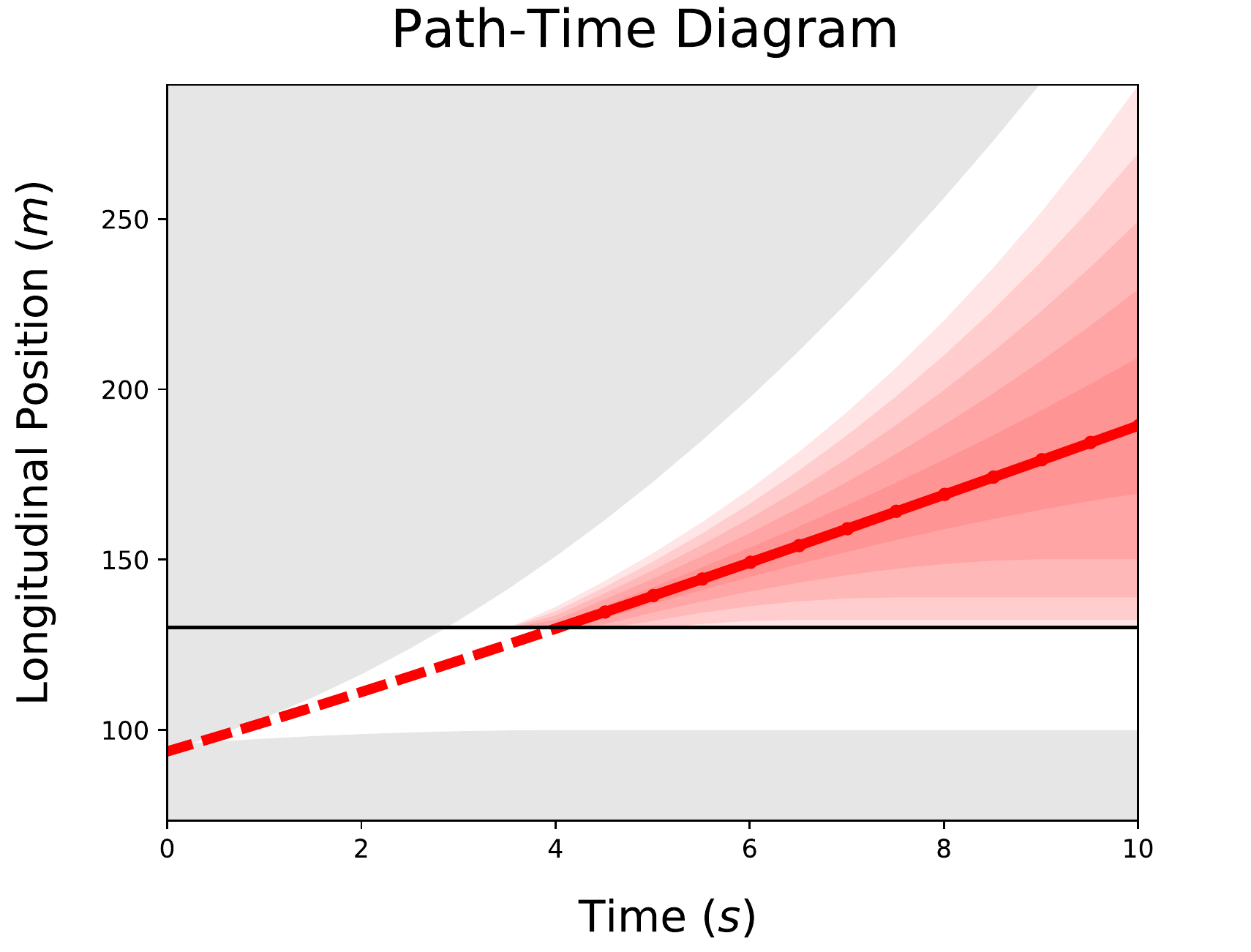}
        \caption{The obtained Gaussian distribution by iteratively calculating the prediction step. 
        The distribution is truncated at the begin of the intersection.}
        \label{fig:figure_02b}
    \end{subfigure}
    
    \caption{Steps of motion prediction of a vehicle. 
    The prediction is unimodal and hence corresponds to a single maneuver hypothesis.}
    \label{fig:figure_02}
\end{figure}

By applying the acceleration values obtained from the IDM
and iteratively calculating the prediction step of the Kalman filter, 
probabilistic prediction is obtained, \textit{cf.\ } Fig~\ref{fig:figure_02}. 
Like this the longitudinal Frenet coordinate for a given point in time $k$ is normal distributed, $x_k \sim \mathcal{N}(\mu_k, \sigma^{2}_{k})$.

\subsection{Prediction of Alternative Maneuvers}
\label{sec:maneuver_prob}

Not only the ego vehicle, but also vehicles in the environment can perform alternative maneuvers. 
The ego vehicle must consider alternative maneuver options which other vehicles might execute 
and assign a probability for each alternative. 
Considering that each of the maneuver hypothesis is Gaussian distributed and is truncated with 
vehicle motion limits or intersection begin, 
the resulting multi-maneuver hypothesis is a truncated Gaussian mixture.

The probability of the alternative maneuvers of other vehicles 
can naively be obtained by comparing the longitudinal component of Frenet coordinates of 
the executed motion with the maneuver hypothesis. 
Other more sophisticated probabilistic classifiers can be used as well.
A dissimilarity measure can for instance be defined as
\begin{equation}
	m(\mathbf{X}_{\mathrm{exec}}, \mathbf{X}_{\mathrm{maneuver}}) = \sum_{k=0}^{n} { ( x_{\mathrm{exec}, n-k} - x_{\mathrm{maneuver}, n-k} ) }^2 \mathrm{,}
\end{equation}
with longitudinal frenet coordinate $x$ and time index $k$ iterating over the past $n + 1$ points in time.

By relative weighting, the maneuver probabilities can be obtained. 
At an intersection, where another vehicle has two maneuver options (\textit{yield}, \textit{drive}), 
the probability that it will yield $p_\mathrm{yield}$ can be given as
\begin{equation}
p_\mathrm{yield} = \frac{m(\mathbf{X}_{\mathrm{exec}}, \mathbf{X}_{\mathrm{yield}})}{m(\mathbf{X}_{\mathrm{exec}}, \mathbf{X}_{\mathrm{yield}}) + m(\mathbf{X}_{\mathrm{exec}}, \mathbf{X}_{\mathrm{drive}}) } ,
\end{equation}
where $\mathbf{X}$ represents the vector filled with trajectory support points.
Note that these relative probabilities correspond to the weights of the components in the Gaussian mixture.

%%%%%%%%%%%%%%%%%%%%%%%%%%%%%%%%%%%%%%%%%%%%%%%%%%%%%%%%%%%%%%%%%%%%%%%%%%%%%%%%%%%%%%%%%%%%%%%%%%%%%%%%%%%%%%%%
\section{Safe Combinatorial Planning}
\label{sec:planning}

In the following, we will first present the fundamentals of our planning method. 
As long as there are no other vehicles around, these formulations apply. 
If there are other vehicles, the uncertainty is nonzero and the planner further considers to 
postpone the combinatorial decision making to a later time. 
The subsequent two subsections describe our modeling and integration into the planning algorithm.
The final subsection describes the selection of the maneuver variant for safe maneuvering. 

\subsection{Fundamentals of the Planner}
\label{sec:planning_fundementals}

Our motion planner is based on the approach presented in \cite{ziegler2014trajectory, ziegler2014making}.
An optimal trajectory minimizes an objective functional of the form
\begin{equation} \label{eq:objective_functional}
J \left(x (t)\right) = \int_{t_0}^{t_0 + T} L (x, \dot{x}, \ddot{x}, \dddot{x}) \, \mathrm{d} t.
\end{equation}
We define $x(t)$ as the position of the vehicle in Frenet frame and 
are subject to \emph{internal} constraints of the vehicle itself and
\emph{external} constraints arising from the environment.

Like presented in \cite{tas2016making} for \texttt{planner lead}, the integrand $L$ is given by
\begin{equation} \label{eq:integrand_continuous}
L = j_\mathrm{v_\mathrm{acc}} + j_\mathrm{v_\mathrm{vel}} + j_\mathrm{v_\mathrm{jrk}} + j_\mathrm{v_\mathrm{coll}} + j_\mathrm{r_\mathrm{vel}} + j_\mathrm{r_\mathrm{acc}} + j_\mathrm{r_\mathrm{jrk}}.
\end{equation}
The summands of integrand consist either of value residual terms $j_\mathrm{v}$ or of range residual terms $j_\mathrm{r}$.
Values residuals serve to approach desired values, 
whereas range residuals serve to limit the parameters to certain interval.
Details are presented already in \cite{tas2016making}.

By utilizing forward finite differences, the integral can be approximated by the sum 
\begin{equation} \label{eq:cost_function}
J^\mathrm{d} (x_0, x_1, \ldots, x_{N-1}) = \sum_{i=0}^{N-4} L (x_i, x_i^\mathrm{d}, x_i^\mathrm{dd}, x_i^\mathrm{ddd} ) 
\end{equation}
which can be minimized by an optimization library. 
We use the solver Ceres \cite{ceres-solver} 
and its automatic differentiation feature for calculating gradients and Hessians.

\subsection{Uncertainty-Aware Planning}
\label{sec:uncertainty_planning}

The summand $ j_\mathrm{v_\mathrm{coll}}$ in Equation~(\ref{eq:integrand_continuous}) 
allows dealing with uncertain environment information. 
The environment model for objects is already presented in Section \ref{sec:environment_model}. 

Given a random variable $Z$ that is Gaussian distributed,
the probability that $Z$ will take values less or equal to $z$ is given by
the cumulative distribution function (cdf)
\begin{equation}
\Phi (z) = \frac{1}{2}+ \frac{1}{2} \operatorname{erf} \left(z/ \sqrt{2}\right) 
\end{equation}
where $\operatorname{erf}$ is the error function and can efficiently be approximated with elementary functions. 

For an object whose predicted motion is given by a truncated Gaussian 
the summand for probabilistic collision avoidance is given by
\begin{equation}
j_\mathrm{v_\mathrm{coll}} = w_{\mathrm{coll}} \, 
\frac
{\Phi \left( {\frac {\mathbf{X}-\mathbf{\mu}}{\sigma }} \right) - 
 \Phi \left( {\frac {\mathbf{a}-\mathbf{\mu}}{\mathbf{\sigma}}} \right)} 
{\Phi \left( {\frac {\mathbf{b}-\mathbf{\mu}}{\mathbf{\sigma}}} \right) - 
 \Phi \left( {\frac {\mathbf{a}-\mathbf{\mu}}{\mathbf{\sigma}}} \right) }
\end{equation}
The factor $w_{\mathrm{coll}}$ serves for weighting of this summand in the multicriteria optimization problem. 
The terms $\mathbf{a}$ and $\mathbf{b}$ are the lower and upper truncation vectors, respectively. 
Further, $\mathbf{\mu}$ and $\mathbf{\sigma}$ are defined as vectors and 
correspond to the mean and standard deviation of the predicted motion.

\subsection{Maneuver-Neutral Combinatorial Planning}
\label{sec:timeshifting_planning}

In some cases during driving, \textit{e.g.\ }at uncontrolled junctions, acceleration lanes etc., 
the drivers execute maneuvers that require interaction with others. 
It is a prerequisite to estimate the intended motion of others. 
As described in \cite{tas2018limited}, we define 
how well the intended maneuver can be perceived by other traffic participants
as the transparency of a planned motion. 
In these cases the drivers may require some time to get a better estimation of other vehicles motion
and consequently perform a \emph{non-transparent} motion. 
A non-transparent motion is typically \emph{maneuver-neutral}, 
in the sense that it is optimal for all of the maneuvers, \emph{e.g.\ }lead or yield.

In the case of where another traffic participant has two possible maneuver alternatives, 
there are two homotopy classes that can be planned.
Instead of planning those two homotopy classes separately, 
we reformulate the optimization problem to optimize both of them. 
Furthermore, to consider postponing the decision making to a later time, we set
the trajectory of homotopy classes identical until some point in time $t_\mathrm{c}$. 
In this way, we reduce the cumulative cost when the other vehicle does not execute the initially predicted maneuver. 
We make sure the postponed time is within a range that is feasible, $t_\mathrm{c} \in [t_{\mathrm{pin}}, t_\mathrm{o})$.
Where, $t_{\mathrm{pin}}$ the last pinned time of the trajectory, 
and $t_\mathrm{o}$ the time at which the mean of the predicted motion of other vehicle crosses the intersection of routes.
Among the alternative maneuvers, the one that crosses the intersection area first is selected as the $t_\mathrm{o}$ value,
\textit{c.f.\ }Fig.~\ref{fig:figure_04}. 
We denote the indices of trajectory support points $x$ that correspond to these times 
simply with "$\mathrm{pin}$" and "$\mathrm{tc}$". 

We reorder the optimization parameters to avoid duplicates and to obtain a single array.
The array consists of three parts: 
First, the parameters until index $\mathrm{tc}$ that are shared by both of the variants. 
Then, the parameters from $\mathrm{tc}$ until the planning horizon for the case of where the ego vehicle leads.
Finally, the parameters for the case at which the ego vehicle yields to the other vehicle.
\begin{equation} \notag	
	\mathbf{X} = 
	(\underbrace{x_{0}, \ldots x_\mathrm{pin}, \ldots x_{\mathrm{tc}}}_{\text{shared}}, \,
	\underbrace{x_{\mathrm{tc}+1}, \ldots x_{N}}_{\text{ego lead}}, \,
	\underbrace{x_{N+1}, \ldots x_{2N - \mathrm{tc}}}_{\text{ego yield}})
\end{equation}
The parts for leading and yielding have the same length and 
each can be linked with the shared part to obtain a proper trajectory of length $N$.
Initialization and constraints have to be given in the same layout. 
We set the profile for initialization \textit{epsilon} distant from the hard maneuver bounds,
which is either the motion limit of the vehicle (in free ride) or the predicted mean position of the other vehicle.

When setting up the different cost terms some extra residuals have to be included.
Using forward differences for velocity we iterate over indices 
$\{0, \ldots, N-1 \}$ and 
$\{N+1, \ldots, 2N - \mathrm{tc} -1 \}$ 
adding all but one residuals presented in Section~\ref{sec:planning_fundementals}.
To connect the shared part with the yield part the addition of a specific residual is needed. 
This is the finite difference between $x_{\mathrm{tc}}$ and $x_{N+1}$.
In a similar fashion, we add two extra residuals for acceleration, and three extras for jerk summands.
The process is done alike for value and range residuals, \textit{cf.\ }Equation \eqref{eq:cost_function}.

With the setup above, both maneuver variants are optimized simultaneously.
And by choosing a suitable value for $\mathrm{tc}$, 
the decision for which one to execute can be delayed.
The setup can also be generalized to handle more than two maneuver variants and multiple other vehicles.

\subsection{Safe Maneuvering in Combinatorial Uncertain Traffic}
\label{sec:safe_planning}

The presented approach is based on continuous replanning. 
As soon as the planning problem is solved and new, up-to-date environment information is available, 
the planning routine is restarted.
As presented in the previous subsection, 
we optimize for the distinct homotopic classes that differ from $t_\mathrm{c}$ on. 

The optimization for different $t_\mathrm{c}$ values can be done in parallel. 
Even though a multitude of alternatives can be calculated when computational power is available,  
we choose only two alternatives: 
$t_\mathrm{c} = t_{\mathrm{pin}}$ and $t_\mathrm{c} = 2 t_{\mathrm{pin}}$.
The first one is identical to the standard planning for two separate trajectories.
In the second case, we consider to postpone the decision making to the next replanning instance 
and drive with a neutral trajectory until then.

Overall there are three alternatives: 
leading, yielding, or driving a neutral trajectory to gather more information.
The decision for one of these is made on the basis of 
the maneuver probabilities of the other vehicle and 
the costs of the planned trajectories,  
in case they satisfy the condition on collision probability. 

The calculation of other vehicle's manuever probabilities is presented in Section~\ref{sec:maneuver_prob}. 
To make a comparison between the probabilities $p_i$, we utilize the information entropy
\begin{equation} 
	\mathrm{H}_{\mathrm{man}} = - \sum_i p_i \log(p_i) .
\end{equation}
If the entropy is below a predefined threshold, 
we assume that the maneuver of the other vehicle is almost certain.
For weight factors imitating normal driving -- where changes in acceleration and jerk are penalized -- 
it is advantageous to do the decison making as soon as possible. 
Therefore,
once the maneuver of the other vehicle is known,
the planner does not further delay the decision and chooses either to lead or to yield.  
A high entropy indicates that the maneuver of the other vehicle is unclear.
In that case it is favourable to delay the decision, perform maneuver neutral driving and gather more information.

When the intention of other vehicle is clear 
and both of the maneuver alternatives have a collision probability less than a threshold, 
the maneuver alternative with yielding the least cost is selected. 
However, the cost calculation is done a different way than presented in Equation~(\ref{eq:integrand_continuous}): 
it includes all the terms but the term for collision probability $ j_\mathrm{v_\mathrm{coll}}$.
The cost arising from collision probability does not reflect the ride quality and is merely 
a safety aspect that needs to be satisfied in any case. 
If the collision probability is higher than the threshold, 
the maneuver is treated as unfeasible and is not considered as an alternative at all. 
In case, both of the manuever variants exceed the collision probability threshold, 
an emergency maneuver -- such as full braking -- can be activated.

%%%%%%%%%%%%%%%%%%%%%%%%%%%%%%%%%%%%%%%%%%%%%%%%%%%%%%%%%%%%%%%%%%%%%%%%%%%%%%%%%%%%%%%%%%%%%%%%%%%%%%%%%%%%%%%%
\section{Experiments}
\label{sec:experiments}

\begin{figure}[h!]
\vspace{2mm}
\includegraphics[width=\columnwidth]{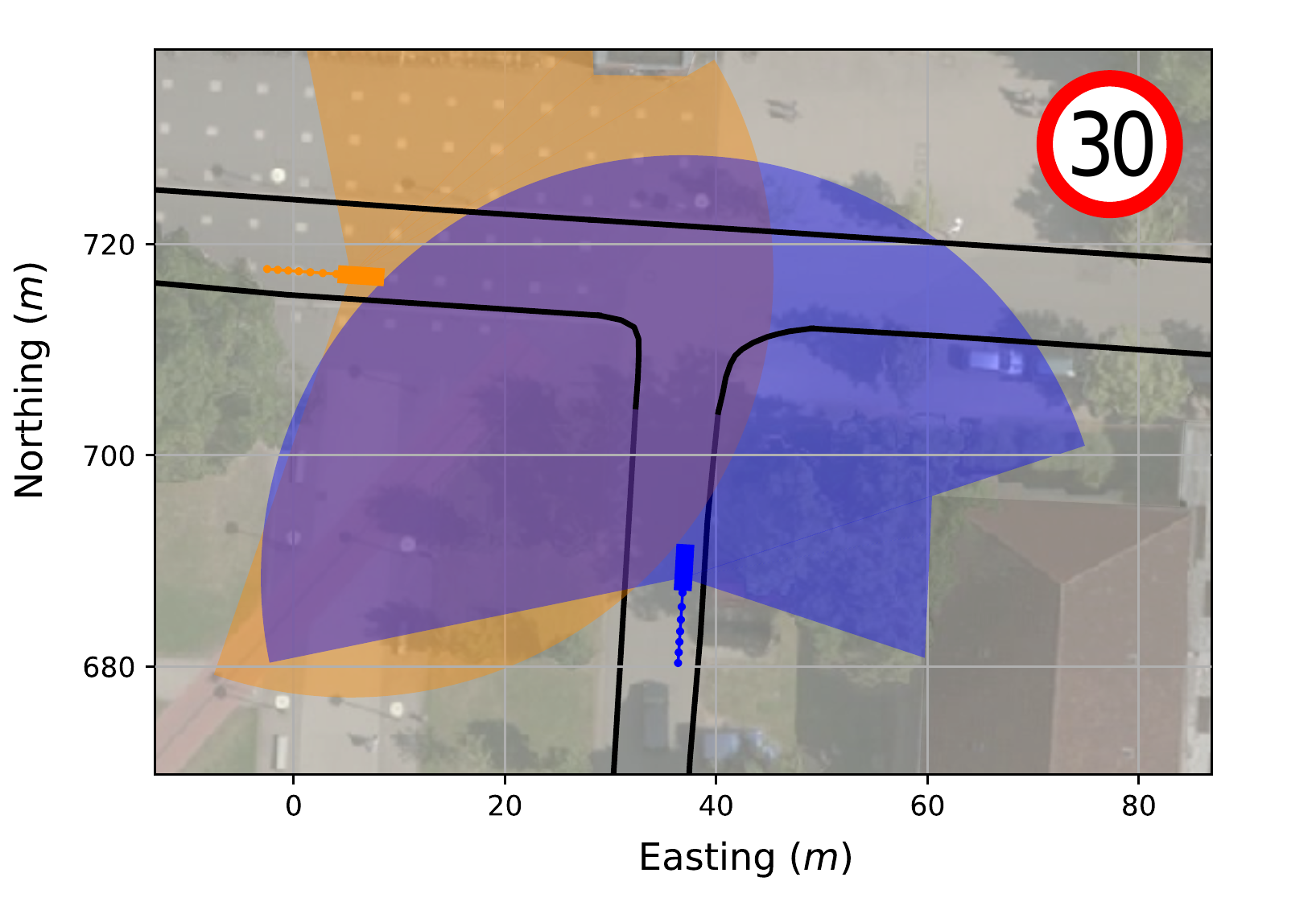}
\caption{Two vehicles approaching an uncontrolled intersection. 
The visible field of views are \ang{210} and have a range of \SI{40}{\metre}. 
After the shown moment maneuver intention estimation, and hence, an interaction will begin.}
\label{fig:figure_03}
\end{figure}

The proposed planner has been tested in closed-loop simulation. 
The use case tested in simulation is shown in Fig.~\ref{fig:figure_03}.
The ego vehicle, depicted with a blue rectangle, approaches an uncontrolled intersection. 
Another vehicle, depicted in orange, is also approaching the same intersection. 
The vehicles can perceive objects, once they are inside the visible field.
At the timestamp shown in the figure, the vehicles are about to detect each other and start to interact.

\begin{figure*}[h!]
\vspace{2mm}
    \centering
    \begin{subfigure}[b]{0.45\textwidth}      
        \includegraphics[width=\textwidth]{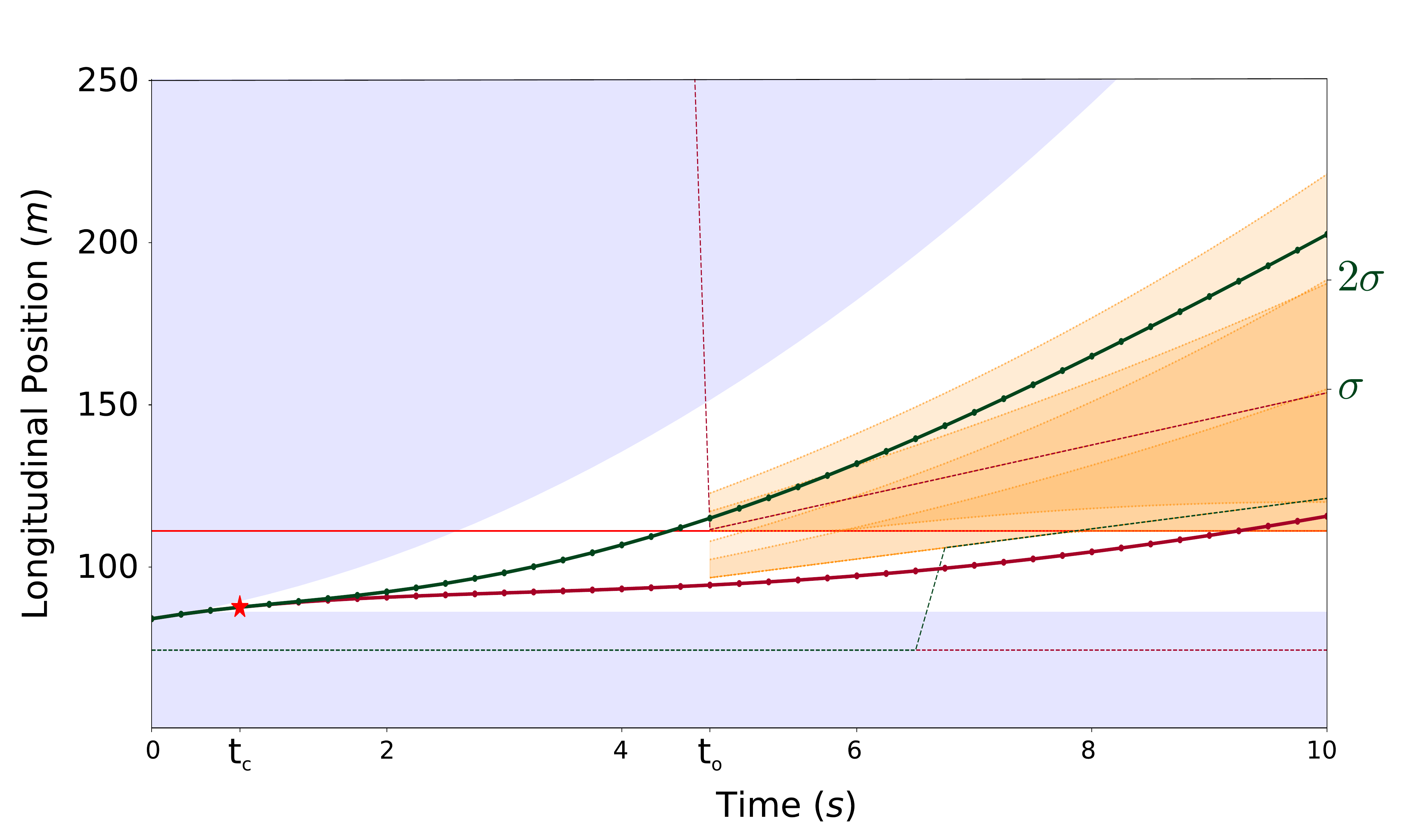}
        \caption{The alternative drive or yield maneuvers at $t_\mathrm{c} = t_\mathrm{pin}$.}
        \label{fig:figure_04a}
    \end{subfigure}
    \begin{subfigure}[b]{0.45\textwidth}      
        \includegraphics[width=\textwidth]{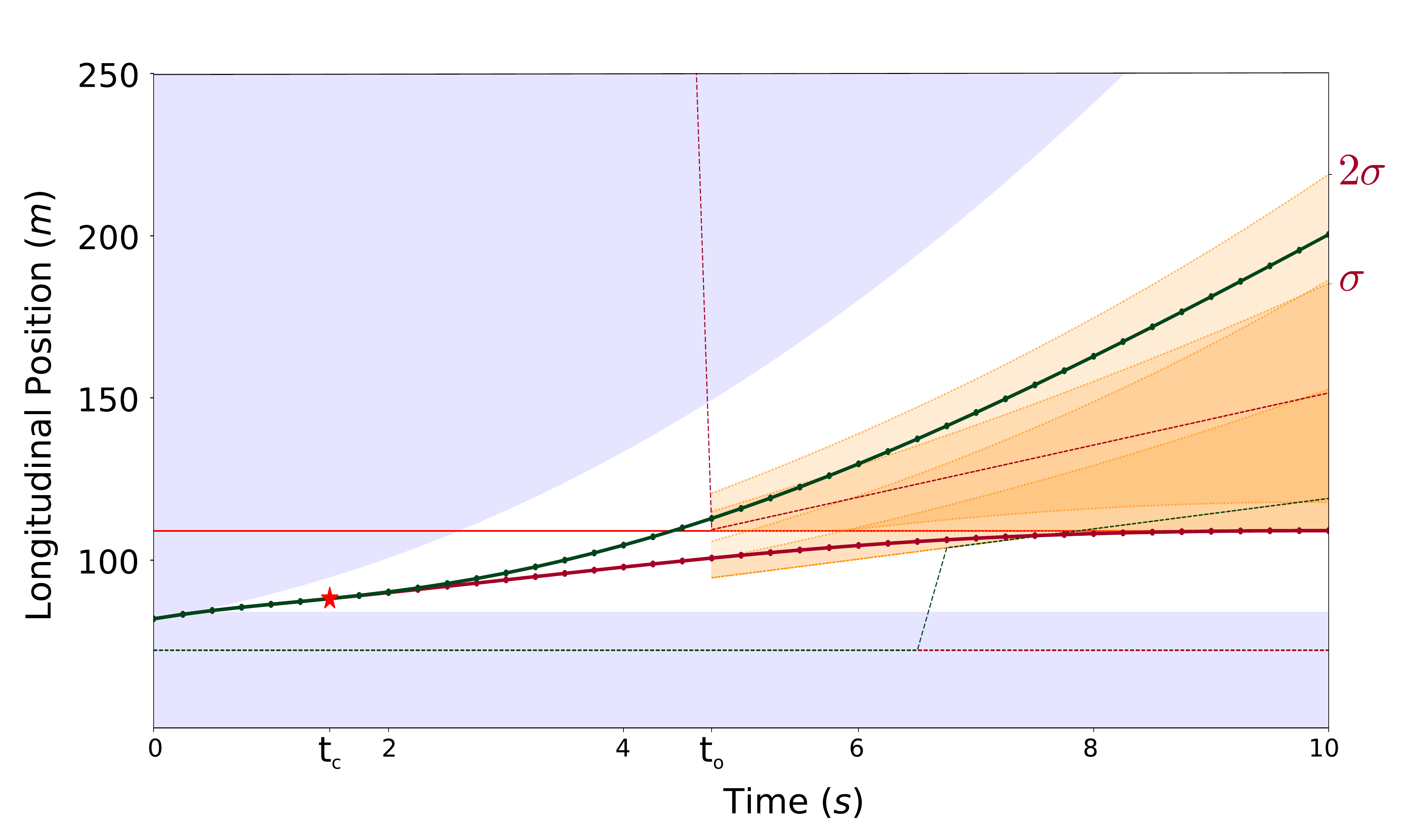}
        \caption{The neutral driving option until next replaning $t_\mathrm{c} = 2t_\mathrm{pin}$.}
        \label{fig:figure_04b}
    \end{subfigure}
\caption{\label{fig:figure_04} 
Planned motion variants on path-time diagram. 
The blue regions are unreachable for the ego vehicle.
The horizontal line corresponds to the merge point of two merging paths. 
With orange, the Gaussian position prediction of the merging vehicle for $2\sigma$ is visualized.
The green and red lines are the maneuver alternatives of the ego vehicle and 
diverge from each other at the star-sign denoted points $t_\mathrm{c}$.
At time $t_\mathrm{o}$ the mean of the predicted motion of the most critical maneuver of the other vehicle crosses the intersection area. 
The distribution of the less critical maneuver of the other vehicle is plotted also below the intersection begin -- the red line -- for better understanding, even though this is not considered in the computation.
}
\end{figure*}

The resulting motion for a timestamp after which the interaction has started 
is presented in Fig.~\ref{fig:figure_04}.
In the use case, the approaching orange vehicle has two maneuver alternatives.
Hence, the position distribution is a truncated Gaussian-mixture. 
In case the ego vehicle leads, 
the orange vehicle has to yield and therefore, 
the Gaussian component at the lower part applies. 
The optimized motion for this case is depicted in green.
The complementary case is depicted in red.

\begin{figure}[h!]
\includegraphics[width=\columnwidth]{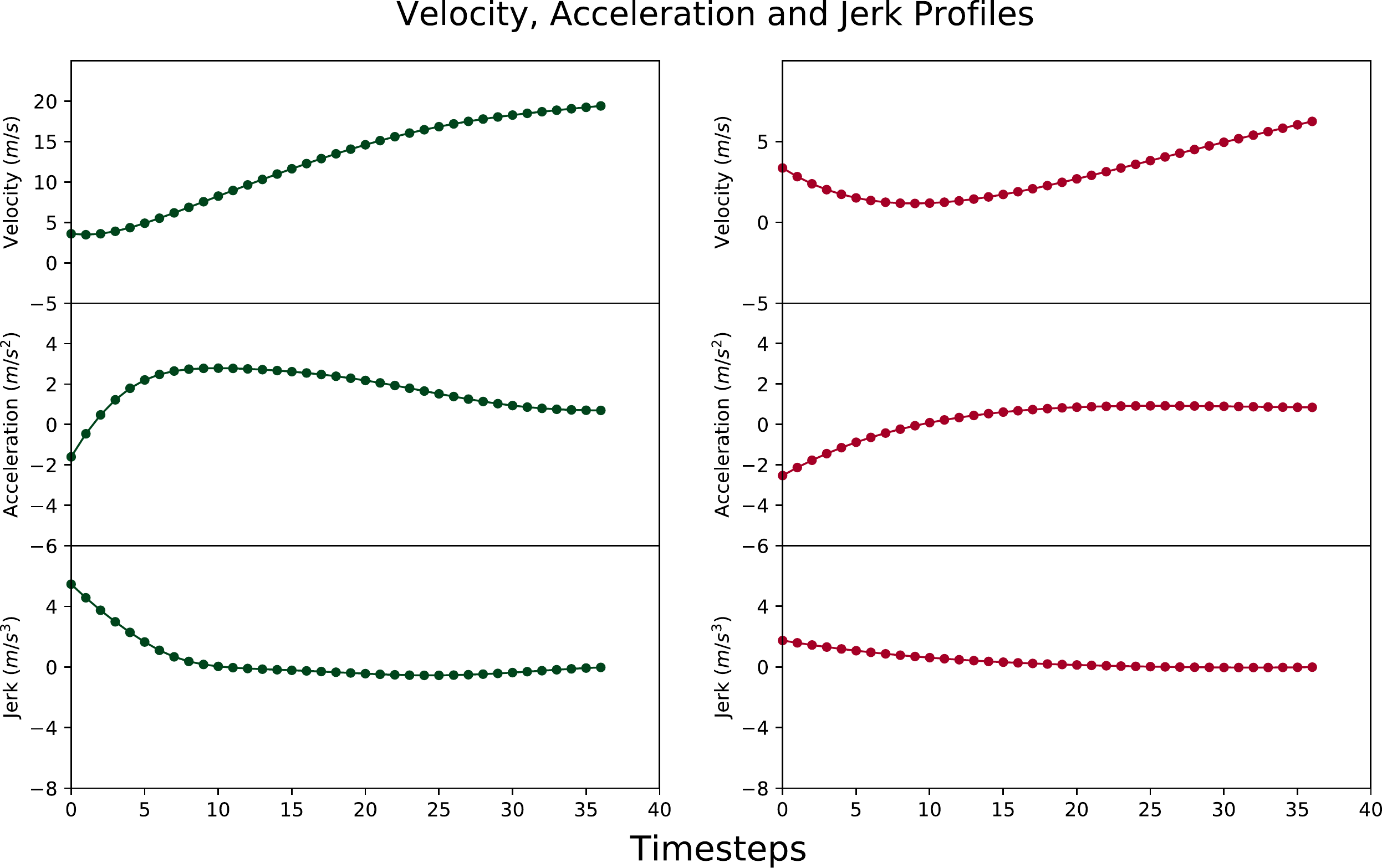}
\caption{The motion profile for both maneuver alternatives for the path 
presented in Fig.~\ref{fig:figure_04a}. 
Values after $t_\mathrm{c}$ are visualized.}
\label{fig:figure_05}
\end{figure}

The velocity, acceleration and jerk components of the planned motion, 
for the both variants in the case $t_\mathrm{c} = t_\mathrm{pin}$ is shown in Fig.~\ref{fig:figure_05}. 
The quality of the motion can easily be inferred from the smoothness of the jerk profile. 

The results for all of the simulated timestamps are given in TABLE~\ref{table:results} in Appendix. %
As shown, the collision probabilities are very low. 
Further, it should be underlined that, 
the highest probability among the support points within the horizon are taken as collision probability and 
these values do not reflect the maneuvers that the ego vehicle can execute to hinder a collision. 
For this reason, these values shall not be interpreted as the \textit{ultimate} collision probabilities. 
A careful analysis highlights the \textit{mostly}\footnote{The obtained final costs depend on the performed number of optimization iterations as well.} 
reduced costs by postponing decision making to a later time 
(compare costs of $t_\mathrm{c} = t_\mathrm{pin}$ and $t_\mathrm{c} = 2 t_\mathrm{pin}$).
The reduction in cost arises from preventing the execution of over-conservative maneuvers. 

In this experiment, 
the number of optimization parameters were $37$ for straight planning, and in the combinatorial case 
$74$ for $t_\mathrm{c} = t_\mathrm{pin}$ and $71$ for $t_\mathrm{c} = 2 t_\mathrm{pin}$. 
The computation time was lower than \SI{50}{\ms} for straight planning and \SI{300}{\ms} for combinatorial planning 
on a Intel Core i7-4710MQ.
These times can be reduced by optimizing the implementation and modifying the solver settings. 
A scalability analysis with the number of feasible combinations is beyond the scope of this paper.

%%%%%%%%%%%%%%%%%%%%%%%%%%%%%%%%%%%%%%%%%%%%%%%%%%%%%%%%%%%%%%%%%%%%%%%%%%%%%%%%%%%%%%%%%%%%%%%%%%%%%%%%%%%%%%%%
\section{Conclusions and Future Work}
\label{sec:conclusion}

Motion planning algortihms, except the ones that are based on MDPs 
need a higher ordered and external module to make combinatorial decisions. 
They typically make combinatorial decisions at the moment which they become feasible and
cannot plan motions that are favorable for all the combinations.
The main contribution of this paper has been twofold:
first, we modeled the effect of uncertain environment information 
on motion planning and considered this in decision making by means of cost terms. 
The second main contribution, which can be identified as the novelty, 
lies in the structure of the planner: 
a local-continuous planner allows intrinsically to consider postponing the combinatorial decision making. 

The simulation results have proven the increased efficiency and safety.
Compared to previous work, the planner considers the effect of uncertainty in
combinatorial decision making. 
The planner switches to the favorable combination only 
when it becomes safe enough.
By doing this, agressive maneuvers leading to a higher cost are prevented.

Our future work will primarily concentrate on a much broader simulation study and a more thorough runtime analysis.
We will further investigate the uncertainties in the prediction, 
such as the effect of estimated IDM-parameters in the predicted maneuvers and motion.
The integration of our previous research on limited visibility and provably safe maneuvering 
\cite{tas2018limited} on the combinatorial decision making will also constitute the baseline of our future research. 
Considering the reaction capabilities of the ego vehicle in planning 
will further allow the planner to insist on the desired maneuver alternative, 
in situations where the other vehicle performs a maneuver that contradicts with our desired maneuver.

%%%%%%%%%%%%%%%%%%%%%%%%%%%%%%%%%%%%%%%%%%%%%%%%%%%%%%%%%%%%%%%%%%%%%%%%%%%%%%%%%%%%%%%%%%%%%%%%%%%%%%%%%%%%%%%%
\section*{Acknowledgements}

The research leading to these results has received funding from the State of Baden-Wuerttemberg in Germany, 
as part of the Tech Center a-Drive project. 
Responsibility for the information and views set out in this publication lies entirely with the authors.

%%%%%%%%%%%%%%%%%%%%%%%%%%%%%%%%%%%%%%%%%%%%%%%%%%%%%%%%%%%%%%%%%%%%%%%%%%%%%%%%%%%%%%%%%%%%%%%%%%%%%%%%%%%%%%%%
\bibliographystyle{IEEEtran}
\bibliography{tas2018decision}

\begin{thebibliography}{10}
\providecommand{\url}[1]{#1}
\csname url@samestyle\endcsname
\providecommand{\newblock}{\relax}
\providecommand{\bibinfo}[2]{#2}
\providecommand{\BIBentrySTDinterwordspacing}{\spaceskip=0pt\relax}
\providecommand{\BIBentryALTinterwordstretchfactor}{4}
\providecommand{\BIBentryALTinterwordspacing}{\spaceskip=\fontdimen2\font plus
\BIBentryALTinterwordstretchfactor\fontdimen3\font minus
  \fontdimen4\font\relax}
\providecommand{\BIBforeignlanguage}[2]{{%
\expandafter\ifx\csname l@#1\endcsname\relax
\typeout{** WARNING: IEEEtran.bst: No hyphenation pattern has been}%
\typeout{** loaded for the language `#1'. Using the pattern for}%
\typeout{** the default language instead.}%
\else
\language=\csname l@#1\endcsname
\fi
#2}}
\providecommand{\BIBdecl}{\relax}
\BIBdecl

\bibitem{tas2014integrating}
\BIBentryALTinterwordspacing
{\"O}.~{\c S}. Ta{\c s}, ``\BIBforeignlanguage{English}{{Integrating
  Combinatorial Reasoning and Continuous Methods for Optimal Motion Planning of
  Autonomous Vehicles}},'' Master's thesis, Karlsruhe Institute of Technology,
  Germany, September 2014. [Online]. Available:
  \url{http://digbib.ubka.uni-karlsruhe.de/volltexte/documents/3370464}
\BIBentrySTDinterwordspacing

\bibitem{bender2015combinatorial}
P.~Bender, {\"O}.~{\c{S}}. Ta{\c{s}}, J.~Ziegler, and C.~Stiller, ``The
  combinatorial aspect of motion planning: Maneuver variants in structured
  environments,'' in \emph{Proc.\ IEEE Intell.\ Veh.\ Symp.}\hskip 1em plus
  0.5em minus 0.4em\relax IEEE, 2015, pp. 1386--1392.

\bibitem{kohlhaas2014semantic}
R.~Kohlhaas, T.~Bittner, T.~Schamm, and J.~M. Z{\"o}llner, ``Semantic state
  space for high-level maneuver planning in structured traffic scenes,'' in
  \emph{Proc.\ IEEE Intell.\ Trans.\ Syst.\ Conf.}, 2014, pp. 1060--1065.

\bibitem{park2015homotopy}
J.~Park, S.~Karumanchi, and K.~Iagnemma, ``Homotopy-based divide-and-conquer
  strategy for optimal trajectory planning via mixed-integer programming,''
  \emph{IEEE Trans. on Robotics}, vol.~31, no.~5, pp. 1101--1115, 2015.

\bibitem{altche2017partitioning}
F.~Altch{\'e} and A.~De~La~Fortelle, ``Partitioning of the free space-time for
  on-road navigation of autonomous ground vehicles,'' in \emph{IEEE Annual
  Conference on Decision and Control (CDC)}, 2017, pp. 2126--2133.

\bibitem{sontges2017computing}
S.~S{\"o}ntges and M.~Althoff, ``Computing the drivable area of autonomous road
  vehicles in dynamic road scenes,'' \emph{IEEE Trans.\ Intell.\ Transp.\
  Syst.}, 2017.

\bibitem{brechtel2014probabilistic}
S.~Brechtel, T.~Gindele, and R.~Dillmann, ``Probabilistic decision-making under
  uncertainty for autonomous driving using continuous pomdps,'' in \emph{Proc.\
  IEEE Intell.\ Trans.\ Syst.\ Conf.}, 2014, pp. 392--399.

\bibitem{hubmann2017decision}
C.~Hubmann, M.~Becker, D.~Althoff, D.~Lenz, and C.~Stiller, ``Decision making
  for autonomous driving considering interaction and uncertain prediction of
  surrounding vehicles,'' in \emph{Proc.\ IEEE Intell.\ Veh.\ Symp.}, 2017, pp.
  1671--1678.

\bibitem{sefati2017towards}
M.~Sefati, J.~Chandiramani, K.~Kreiskoether, A.~Kampker, and S.~Baldi,
  ``Towards tactical behaviour planning under uncertainties for automated
  vehicles in urban scenarios,'' in \emph{Proc.\ IEEE Intell.\ Trans.\ Syst.\
  Conf.}, 2017, pp. 1--7.

\bibitem{gritschneder2016adaptive}
F.~Gritschneder, P.~Hatzelmann, M.~Thom, F.~Kunz, and K.~Dietmayer, ``Adaptive
  learning based on guided exploration for decision making at roundabouts,'' in
  \emph{Proc.\ IEEE Intell.\ Veh.\ Symp.}, 2016, pp. 433--440.

\bibitem{hoermann2017entering}
S.~Hoermann, F.~Kunz, D.~Nuss, S.~Renter, and K.~Dietmayer, ``Entering
  crossroads with blind corners. a safe strategy for autonomous vehicles,'' in
  \emph{Proc.\ IEEE Intell.\ Veh.\ Symp.}, 2017, pp. 727--732.

\bibitem{lefevre2014survey}
S.~Lef{\`e}vre, D.~Vasquez, and C.~Laugier, ``A survey on motion prediction and
  risk assessment for intelligent vehicles,'' \emph{Robomech Journal}, vol.~1,
  no.~1, p.~1, 2014.

\bibitem{treiber2000congested}
M.~Treiber, A.~Hennecke, and D.~Helbing, ``Congested traffic states in
  empirical observations and microscopic simulations,'' \emph{Physical review
  E}, vol.~62, no.~2, p. 1805, 2000.

\bibitem{xu2014motion}
W.~Xu, J.~Pan, J.~Wei, and J.~M. Dolan, ``Motion planning under uncertainty for
  on-road autonomous driving,'' in \emph{IEEE Int. Conf. on Robotics and
  Automation (ICRA)}, 2014, pp. 2507--2512.

\bibitem{ziegler2014trajectory}
J.~Ziegler, P.~Bender, T.~Dang, and C.~Stiller, ``{Trajectory planning for
  Bertha — a local, continuous method},'' in \emph{Proc.\ IEEE Intell.\ Veh.\
  Symp.}, 2014, pp. 450--457.

\bibitem{ziegler2014making}
J.~Ziegler, P.~Bender, M.~Schreiber, H.~Lategahn, T.~Strauss, C.~Stiller,
  T.~Dang, U.~Franke, N.~Appenrodt, C.~G. Keller \emph{et~al.}, ``Making
  {B}ertha drive — {A}n autonomous journey on a historic route,'' \emph{IEEE
  Intell.\ Transp.\ Syst.\ Mag.}, vol.~6, no.~2, pp. 8--20, 2014.

\bibitem{tas2016making}
{\"O}.~{\c S}. Ta{\c s}, N.~O. Salscheider, F.~Poggenhans, S.~Wirges,
  C.~Bandera, M.~R. Zofka, T.~Strauss, J.~M. Z{\"o}llner, and C.~Stiller,
  ``{Making Bertha Cooperate - Team AnnieWAY's Entry to the 2016 Grand
  Cooperative Driving Challenge},'' \emph{IEEE Trans.\ Intell.\ Transp.\
  Syst.}, vol.~19, no.~4, pp. 1262--1276, April 2018, {Date of Publication: 02
  October 2017}.

\bibitem{ceres-solver}
S.~Agarwal, K.~Mierle, and Others, ``Ceres solver,''
  \texttt{http://ceres-solver.org}.

\bibitem{tas2018limited}
{\"O}.~{\c S}. Ta{\c s} and C.~Stiller, ``{Limited Visibility and Uncertainty
  Awware Motion Planning for Automated Driving},'' in \emph{Proc.\ IEEE
  Intell.\ Veh.\ Symp.}, Changshu, China, June 2018.

\end{thebibliography}

%%%%%%%%%%%%%%%%%%%%%%%%%%%%%%%%%%%%%%%%%%%%%%%%%%%%%%%%%%%%%%%%%%%%%%%%%%%%%%%%%%%%%%%%%%%%%%%%%%%%%%%%%%%%%%%%
\section*{Appendix}
\label{sec:appendix}

\begin{threeparttable}[htb]
\caption{Results of the maneuver predictions and optimized motion profiles throughout the simulation.}
\label{table:results}
\small
\setlength\tabcolsep{0pt}
\begin{tabular*}{\linewidth}{@{\extracolsep{\fill}} c l c l c c c r @{}}
\toprule
Timestamp & & Alternative & & Collision Prob. & & & Cost \\
\midrule
75.75     & & -           & & -               & & & 18.881  \\[1ex]
76.00     & & -           & & -               & & & 19.005  \\[1ex]
76.25     & & -           & & -               & & & 19.174  \\[1ex]
                        & & lead ($t_\mathrm{pin} $)    &          & 0.011 & & & 114.674  \\
                        & & lead ($2t_\mathrm{pin} $)   &          & 0.010 & & & 135.239  \\
\raisebox{0.5ex}{76.50} & & follow ($t_\mathrm{pin} $)  &          & 0.033 & & & 112.568  \\
                        & & follow ($2t_\mathrm{pin} $) &          & 0.036 & & & 67.143   \\[1ex]
                        & & lead ($t_\mathrm{pin} $)    &          & 0.011 & & & 108.144  \\
                        & & lead ($2t_\mathrm{pin} $)   &          & 0.010 & & & 87.775   \\
\raisebox{0.5ex}{76.75} & & follow ($t_\mathrm{pin} $)  &          & 0.030 & & & 105.737  \\
                        & & follow ($2t_\mathrm{pin} $) &          & 0.033 & & & 68.458   \\[1ex]
                        & & lead ($t_\mathrm{pin} $)    &          & 0.010 & & & 69.069  \\
                        & & lead ($2t_\mathrm{pin} $)   &          & 0.010 & & & 87.775  \\
\raisebox{0.5ex}{77.00} & & follow ($t_\mathrm{pin} $)  &          & 0.026 & & & 67.711  \\
                        & & follow ($2t_\mathrm{pin} $) &          & 0.029 & & & 27.770  \\[1ex]
                        & & lead ($t_\mathrm{pin} $)    &          & 0.009 & & & 66.890  \\
                        & & lead ($2t_\mathrm{pin} $)   &          & 0.009 & & & 84.518  \\
\raisebox{0.5ex}{77.25} & & follow ($t_\mathrm{pin} $)  &          & 0.020 & & & 65.782  \\
                        & & follow ($2t_\mathrm{pin} $) &          & 0.022 & & & 28.771  \\[1ex]
\bottomrule
\end{tabular*}
\end{threeparttable}
\end{document}